\documentclass[10pt,a4paper,twocolumn]{article}
\usepackage[a4paper,left=1.45cm,right=1.45cm,top=1.6cm,bottom=1.6cm,columnsep=0.36cm]{geometry}
\usepackage{times}
\usepackage{graphicx}
\usepackage{amsmath,amssymb}
\usepackage{booktabs}
\usepackage{caption}
\usepackage{titlesec}
\usepackage{fancyhdr}
\usepackage{enumitem}
\usepackage{xcolor}
\usepackage{tikz}
\usetikzlibrary{arrows.meta,positioning}

\titleformat{\section}{\bfseries\fontsize{10.8}{11.8}\selectfont}{\thesection}{0.45em}{}
\titleformat{\subsection}{\itshape\fontsize{9.6}{10.5}\selectfont}{\thesubsection}{0.45em}{}
\titlespacing*{\section}{0pt}{0.40em}{0.16em}
\titlespacing*{\subsection}{0pt}{0.30em}{0.10em}
\setlist[itemize]{leftmargin=1.0em,itemsep=0pt,topsep=1pt,parsep=0pt}

\begin{document}
\setlength{\abovedisplayskip}{3pt}\setlength{\belowdisplayskip}{3pt}
\setlength{\abovedisplayshortskip}{2pt}\setlength{\belowdisplayshortskip}{2pt}
\twocolumn[
\begin{center}
{\fontsize{15.3}{17}\selectfont\bfseries
Generalizing Geometry-Guided Mamba as a Plug-and-Play Context Module for CNN-based Semantic Segmentation
\par}
\vspace{0.30em}
Hsin-Jui Pan$^{1}$, Sheng-Wei Chan$^{1}$, Chun-Po Shen$^{1}$, Jen-Shiun Chiang$^{1*}$\\
{\small $^{1}$Department of Electrical and Computer Engineering, Tamkang University, New Taipei City, Taiwan\\
$^{*}$chiang@mail.tku.edu.tw}
\vspace{0.40em}
\end{center}

\vspace{0.35em}

\noindent\textbf{Abstract}---CNN-based semantic segmentation networks usually rely on context heads such as ASPP, PPM, or attention modules to enlarge the receptive field. These heads are effective but may introduce heavy computation, memory cost, or boundary leakage. This paper revisits Directional Geometric Mamba (G-Mamba) from DGM-Net and studies it as a plug-and-play context aggregation module rather than a completely new segmentation architecture. The key idea is to inject geometric guidance into the selective scan process, allowing long-range feature propagation to be modulated by boundary and centripetal-flow cues. We replace the original context heads of six representative CNN segmentation models, including DeepLabV3+, DANet, CCNet, PSPNet, PSANet, and OCRNet, while keeping the ResNet-101 backbone unchanged. On CCNet, we additionally compare serial and parallel combinations of criss-cross attention and the G-Mamba block, with the parallel head performing best. Results on Cityscapes show consistent mIoU gains with only moderate extra GFLOPs at $1024\times1024$ resolution, suggesting that geometry-guided SSM modules can serve as practical alternatives or enhancements to conventional CNN context heads.
\vspace{0.55em}
]

\section{Introduction}
Semantic segmentation assigns a semantic category to every image pixel and requires both local boundary precision and global scene understanding. CNN-based methods such as FCN, DeepLabV3+, and PSPNet provide strong local feature extraction and efficient dense prediction, but their context reasoning is usually handled by additional modules such as atrous spatial pyramid pooling (ASPP), pyramid pooling (PPM), or attention-based heads \cite{long2015fully,chen2018encoder,zhao2017pyramid}. Although these modules improve the effective receptive field, they depend on either multi-branch convolutional sampling or pairwise spatial interactions, both of which become costly for high-resolution segmentation. State Space Models (SSMs), especially Mamba, provide an attractive alternative because long-range dependency modeling can be performed with linear complexity \cite{gu2023mamba}. Recent visual SSMs such as VMamba and Vim further show that selective scan operations can be adapted to vision tasks \cite{liu2024vmamba,zhu2024vision}. However, standard visual scans are often isotropic: pixels are propagated along fixed routes rather than object geometry, which blurs thin structures and leaks semantics across boundaries. This work follows the observation in DGM-Net that geometry can steer SSM propagation \cite{chan2026breaking}: geometric priors derived from CNN features inject the boundary and structure awareness that a plain Mamba scan lacks. Instead of reusing the whole DGM-Net architecture, we decouple G-Mamba and ask a narrower question: can the geometry-guided Mamba block replace or enhance the context modules of existing CNN segmentation networks? The backbone, crop size, output stride, and training setting are fixed; only the context head is replaced (Fig.~\ref{fig:method}(a)), so the measured gain is directly attributable to the context module. On CCNet, we further compare serial and parallel combinations of criss-cross attention and the G-Mamba block (Fig.~\ref{fig:method}(c)).

\begin{figure}[!t]
\centering
\begin{tikzpicture}[
  every node/.style={font=\scriptsize, align=center, inner sep=1.6pt},
  blk/.style={draw=gray!60!black, rounded corners=2pt, minimum height=4.2mm, minimum width=1.0cm, line width=0.45pt, fill=white},
  blkb/.style={draw=blue!55!black, rounded corners=2pt, minimum height=4.2mm, line width=0.45pt, fill=blue!8},
  blkhi/.style={draw=red!75!black, rounded corners=2pt, minimum height=4.2mm, line width=0.8pt, fill=red!15, font=\scriptsize\bfseries},
  blkcs/.style={draw=blue!55!black, rounded corners=1.5pt, minimum height=3.4mm, minimum width=0.34cm, line width=0.45pt, fill=blue!8, font=\tiny, inner sep=0.5pt},
  blkms/.style={draw=red!75!black, rounded corners=1.5pt, minimum height=3.4mm, minimum width=0.34cm, line width=0.7pt, fill=red!15, font=\tiny\bfseries, inner sep=0.5pt},
  arr/.style={-{Stealth[length=1.4mm,width=1.2mm]}, line width=0.5pt, draw=gray!55!black},
  arrhi/.style={-{Stealth[length=1.4mm,width=1.2mm]}, line width=0.65pt, draw=red!75!black, dashed},
]
% ===== (a) plug-and-play replacement =====
\node[font=\bfseries\tiny] at (-0.55,0.47) {(a)};
\node[blk, minimum width=0.85cm] (in)  at (-0.15,0) {Input};
\node[blk, minimum width=1.5cm]  (bb)  at (1.35,0) {ResNet-101};
\node[blkhi, minimum width=1.55cm] (gm) at (3.45,0) {G-Mamba};
\node[blk, minimum width=1.25cm] (dec) at (5.45,0) {Decoder};
\node[blk, minimum width=1.15cm] (out) at (6.95,0) {Output};
\draw[arr] (in)--(bb); \draw[arr] (bb)--(gm); \draw[arr] (gm)--(dec); \draw[arr] (dec)--(out);
\draw[gray!55!black, dashed, rounded corners=2pt, line width=0.4pt] (2.50,-0.30) rectangle (4.40,0.30);
\node[font=\tiny, text=gray!45!black] at (3.45,0.47) {context-head slot};
\node[blk, draw=gray!45, text=gray!35!black, fill=gray!4, font=\tiny] (orig) at (3.45,-0.92) {ASPP\,/\,PPM\,/\,DA\,/\,CC\,/\,PSA\,/\,OCR};
\draw[arrhi] (orig.north) -- (gm.south) node[midway, right, font=\tiny, text=red!70!black] {replace};

% ===== (b) G-Mamba context module: explicit two-path dataflow =====
\node[font=\bfseries\tiny] at (-0.60,-2.05) {(b)};
% F at far left, vertically centered between the two rows
\node[blk, minimum width=0.55cm] (f) at (-0.05,-2.05) {$F$};
% split dot just right of F
\coordinate (sp) at (0.55,-2.05);
\draw[line width=0.5pt, draw=gray!55!black] (f.east) -- (sp);
\fill[gray!55!black] (sp) circle (0.5pt);
% --- upper branch: priors -> D ---
\node[blkb, minimum width=1.6cm]  (pr) at (2.30,-1.50) {priors $V,\Phi,D_c$};
\node[blkb, minimum width=1.85cm] (dd) at (5.05,-1.50) {$D{=}\sigma(D_c{+}\Delta D)$};
\draw[arr] (sp) |- (pr.west);
\draw[arr] (pr) -- (dd);
% --- lower row: T_s -> (mul) -> scan ---
\node[blkb, minimum width=2.35cm] (ts) at (2.05,-2.60) {$T_s{=}1{+}D\cdot\mathrm{ReLU}(\Phi_{dir}^{s})$};
\node[draw=gray!60!black, circle, inner sep=0.8pt, line width=0.5pt, fill=white] (mul) at (4.30,-2.60) {$\scriptstyle\odot$};
\node[blkhi, minimum width=1.6cm] (scan) at (6.10,-2.60) {selective scan};
% D comes down into T_s
\draw[arr] (dd.south) -- (5.05,-2.05) -| (ts.north);
% direct branch: F drops down and enters the product from below (the other operand)
\draw[arr] (sp) -- (0.55,-3.28) -| (mul.south);
% T_s -> mul (from the left) -> scan
\draw[arr] (ts.east) -- (mul.west);
\draw[arr] (mul.east) -- (scan.west);
% label the product as F ⊙ T_s
\node[font=\tiny, text=gray!45!black] at (4.30,-2.30) {$F\odot T_s$};
% Cascade note as a right-aligned inset, clearly separate
\node[font=\tiny, text=gray!45!black, anchor=east] at (7.05,-3.70) {Cascade: $(N{-}1)$ plain scans $\rightarrow$ final geometry-guided scan};

% ===== (c) hybrid compositions on the CCNet host =====
\node[font=\bfseries\tiny] at (-0.60,-4.28) {(c)};
% serial: CCA -> G-Mamba
\node[blkcs, minimum width=0.55cm] (s1a) at (0.62,-4.28) {CCA};
\node[blkms, minimum width=1.00cm] (s1b) at (1.75,-4.28) {G-Mamba};
\draw[arr] (s1a)--(s1b);
% serial: G-Mamba -> CCA
\node[blkms, minimum width=1.00cm] (s2a) at (3.15,-4.28) {G-Mamba};
\node[blkcs, minimum width=0.55cm] (s2b) at (4.28,-4.28) {CCA};
\draw[arr] (s2a)--(s2b);
% parallel: CCA and G-Mamba on the same input, outputs fused
\coordinate (pin) at (4.95,-4.28);
\draw[line width=0.5pt, draw=gray!55!black] (4.75,-4.28) -- (pin);
\fill[gray!55!black] (pin) circle (0.5pt);
\node[blkcs, minimum width=0.55cm] (pa) at (5.85,-4.06) {CCA};
\node[blkms, minimum width=1.00cm] (pb) at (5.85,-4.50) {G-Mamba};
\node[font=\scriptsize, inner sep=0.2pt] (jn) at (6.90,-4.28) {$\oplus$};
\draw[arr] (pin) |- (pa.west);
\draw[arr] (pin) |- (pb.west);
\draw[arr] (pa.east) -| (jn.north);
\draw[arr] (pb.east) -| (jn.south);
\draw[arr] (jn.east) -- (7.28,-4.28);
\end{tikzpicture}
\caption{(a) Plug-and-play protocol: only the context head of each host network is replaced by G-Mamba or Cascade G-Mamba; the ResNet-101 backbone, decoder, and training setting are unchanged. (b) The G-Mamba context module: geometric priors ($V$, $\Phi$, $D_c$) predicted from $F$ form the detail prompt $D$ and the spatial prompt $T_s$, and the reweighted feature $F\odot T_s$ enters the selective scan. In Cascade G-Mamba, plain multi-directional scans build macro context before the final geometry-guided scan. (c) Hybrid compositions on the CCNet host: criss-cross attention (CCA, blue) and the G-Mamba block (red), combined serially in either order or in parallel.}
\label{fig:method}
\end{figure}

\section{Geometry-Guided Context Module}
\subsection{Geometric prompt construction}
Given an intermediate feature map $F\in\mathbb{R}^{C\times H\times W}$, G-Mamba first predicts lightweight geometric priors from the CNN features (Fig.~\ref{fig:method}(b)). Following the DGM-Net design, the module uses a centripetal potential map $V$, a directional flow field $\Phi$, and a coarse morphological boundary map $D_c$ \cite{chan2026breaking}. The potential map describes the tendency toward object centers, while the boundary map emphasizes thin or high-frequency structures such as poles, fences, and object contours. The flow field is obtained as a first-order directional response of the potential map and provides scan-aware orientation. To avoid an overly noisy boundary prior, a residual refinement branch predicts $\Delta D$ from the context feature, producing the final detail prompt
\begin{equation}
D = \sigma(D_c + \Delta D),
\end{equation}
where $\sigma(\cdot)$ denotes the sigmoid function. For a scan direction $s$, the flow response is projected to the scan axis as $\Phi_{dir}^{s}$. The spatial prompt is then computed as
\begin{equation}
T_s = 1 + D \cdot \mathrm{ReLU}(\Phi_{dir}^{s}),
\end{equation}
and the input to the selective scan is reweighted by
\begin{equation}
F_s' = F \odot T_s.
\end{equation}
Here $\odot$ denotes element-wise multiplication. This formulation keeps the linear scan mechanism of Mamba but changes the scan intensity according to structural cues. Regions near boundaries or along meaningful centripetal directions receive stronger recurrent propagation, while irrelevant background diffusion is suppressed.

\subsection{Cascade G-Mamba}\label{ssec:cascade}
A single G-Mamba layer provides geometry-aware propagation, but it may still be sensitive to local noise if the global semantic field is not sufficiently stable. Therefore, we also test Cascade G-Mamba. The first stages use standard multi-directional Mamba scanning to build a macro semantic context. The final stage applies the geometric prompt above as a precision filter. This design separates the roles of ``what'' and ``where'': early scans aggregate object-level context, while the geometry-guided scan sharpens boundary-sensitive regions. Unlike ASPP or PPM, this context aggregation is not based on fixed dilation rates or pooling bins; the effective receptive field is dynamically shaped by $D$ and $\Phi$.

\section{Plug-and-Play Replacement Protocol}
To evaluate generality, G-Mamba is inserted into six CNN segmentation architectures by replacing only the original context head: ASPP in DeepLabV3+, PPM in PSPNet, dual attention in DANet, criss-cross attention in CCNet, point-wise spatial attention in PSANet, and the object-contextual aggregation head in OCRNet \cite{fu2019dual,huang2019ccnet,zhao2018psanet,yuan2020object}. The ResNet-101 backbone \cite{he2016deep} is kept unchanged for all models.

\section{Experiments}
\subsection{Setup}
All experiments are evaluated on Cityscapes \cite{cordts2016cityscapes}. Each model uses a ResNet-101 backbone with output stride 16, crop size $768\times768$, and batch size 4 on a single RTX 3060 Ti GPU with 8GB memory. The same training protocol is applied to the original baseline, +G-Mamba, and +Cascade G-Mamba variants.

\subsection{Accuracy comparison}
Table~\ref{tab:miou} summarizes the replacement results. G-Mamba improves all six baselines, and Cascade G-Mamba further improves the results in every tested case. The gains are especially clear for DANet and CCNet, where replacing dense attention-style context with geometry-guided sequence modeling improves mIoU by more than two points. This suggests that the module is not only useful for ASPP-like convolutional heads, but also for attention-based context aggregation.

\begin{table}[t]
\centering
\caption{Cityscapes validation mIoU comparison. All models use ResNet-101, OS=16, crop $768\times768$, and batch size 4.}
\label{tab:miou}
\footnotesize
\setlength{\tabcolsep}{7pt}
\begin{tabular}{lccc}
\toprule
Architecture & Baseline & +G-Mamba & +Cascade \\
\midrule
DeepLabV3+ & 75.17 & 76.58 & \textbf{77.57} \\
DANet      & 76.32 & 78.49 & \textbf{79.16} \\
CCNet      & 75.54 & 77.83 & \textbf{78.81} \\
PSPNet     & 77.52 & 78.76 & \textbf{79.42} \\
PSANet     & 76.79 & 77.02 & \textbf{77.23} \\
OCRNet     & 78.22 & 79.01 & \textbf{79.57} \\
\bottomrule
\end{tabular}
\end{table}

\subsection{Hybrid composition on CCNet}
Since the largest gains in Table~\ref{tab:miou} appear on the attention-based hosts, we further ask how criss-cross attention (CCA) and the G-Mamba block should be combined on CCNet: serially in either order, or in parallel (Fig.~\ref{fig:method}(c)). As shown in Table~\ref{tab:order}, the two serial orders differ sharply: CCA$\,\rightarrow\,$G-Mamba reaches 78.80 mIoU, essentially matching Cascade G-Mamba (78.81), whereas the reverse order reaches only 76.15 (+0.61 over the baseline), likely because the subsequent attention passes re-mix the boundary-sensitive output of the scan along axial routes. This asymmetry supports the ``what-then-where'' design of Section~\ref{ssec:cascade}: the geometry-guided scan works best as a final precision filter on features that already carry stable context, whether that context is built by attention or by plain scans. The parallel head performs best with 79.20 mIoU (+3.66), suggesting that the two mechanisms extract complementary context and that fusing them outperforms conditioning one on the other.
% TODO: parallel 分支的融合方式（相加或 concat+conv）請確認；圖 (c) 目前以 $\oplus$ 表示相加。

\begin{table}[t]
\centering
\caption{Hybrid composition ablation on the CCNet host (Cityscapes val mIoU, same setting as Table~\ref{tab:miou}).}
\label{tab:order}
\footnotesize
\setlength{\tabcolsep}{7pt}
\begin{tabular}{lc}
\toprule
Structure & mIoU \\
\midrule
CCA (CCNet baseline) & 75.54 \\
CCA $\rightarrow$ G-Mamba & 78.80 \\
G-Mamba $\rightarrow$ CCA & 76.15 \\
CCA $\parallel$ G-Mamba (parallel) & \textbf{79.20} \\
\bottomrule
\end{tabular}
\end{table}

\subsection{Computation at high resolution}
As shown in Table~\ref{tab:gflops}, at the $1024\times1024$ input resolution adding G-Mamba increases the computation only moderately. The increase is between 33 and 51 GFLOPs, i.e., 2.6--5.0\% over the original models, so the overhead is relatively small while the accuracy gain in Table~\ref{tab:miou} remains consistent.

\begin{table}[t]
\centering
\caption{GFLOPs comparison at $1024\times1024$ resolution. PSANet is omitted because it exceeds the 8GB GPU memory at this resolution.}
\label{tab:gflops}
\footnotesize
\setlength{\tabcolsep}{6pt}
\begin{tabular}{lcccc}
\toprule
Architecture & Original & +G-Mamba & $\Delta$ & $\Delta$ (\%) \\
\midrule
DeepLabV3+ & 1018.0 & 1065.0 & +47.0 & +4.6 \\
DANet      & 1272.0 & 1306.0 & +34.0 & +2.7 \\
CCNet      & 1103.0 & 1154.0 & +51.0 & +4.6 \\
PSPNet     & 1026.0 & 1077.0 & +51.0 & +5.0 \\
OCRNet     & 1246.0 & 1279.0 & +33.0 & +2.6 \\
\bottomrule
\end{tabular}
\end{table}

\section{Discussion and Conclusion}
The results indicate that G-Mamba can be treated as a transferable context aggregation block rather than being restricted to its original DGM-Net architecture. The main benefit comes from combining two complementary biases: CNN features provide local texture and boundary evidence, while Mamba provides efficient long-range propagation. The geometric prompt connects these two parts by guiding scan intensity with boundary and centripetal-flow information. This short study thus provides evidence that geometry-guided SSMs are practical drop-in replacements for common CNN context heads. The CCNet ablation further shows that composition matters: the serial CCA$\,\rightarrow\,$G-Mamba head is on par with Cascade G-Mamba, while the parallel head performs best (79.20 mIoU).
Future work will include parameter count, latency, and memory profiling, as well as broader evaluation on ADE20K and additional lightweight backbones.

\section*{Acknowledgements}
This research work is partially supported by National Science and Technology Council, Taiwan, under grant number 114-2221-E-032-011.

\end{document}